%% file: paper.tex
\documentclass[]{article}
\usepackage{proceed2e}
\usepackage{enumerate}
\usepackage{algpseudocode}
\usepackage[ruled]{algorithm}
\usepackage{url}
\usepackage{framed}
\usepackage{amsfonts,amsmath,amsthm,amssymb}
\usepackage{graphicx}
\usepackage{url}
\usepackage{color}
\usepackage[round]{natbib}
\usepackage{comment}

\newcommand {\IE} {\ensuremath {\mathbb{E}}}

\newcommand {\term}[1] {\textbf{#1}}

\newcommand {\Evidence} {\mathcal{E}}
\newcommand {\R} {\mathbb{R}}

\newcommand {\given} {\mid} %{\mathrel{|}}
\newcommand {\sample} {\sim}
\newcommand {\iid} {\stackrel{\rm iid}{\sample}}
\newcommand {\Nat} {\mathbb{N}}
\DeclareMathOperator*{\argmax}{argmax}

\newtheorem{thm}{Theorem}
\newtheorem{dfn}[thm]{Definition}

\newtheorem{crl}[thm]{Corollary}
\newtheorem{example}{Example}

\newcommand{\secref}[1]{Section~\ref{#1}}
\newcommand{\secrefs}[2]{Sections~\ref{#1} and~\ref{#2}}
\newcommand{\figref}[1]{Figure~\ref{#1}}
\renewcommand{\eqref}[1]{Equation~(\ref{#1})}
\newcommand{\eqrefs}[2]{Equations~(\ref{#1}) and~(\ref{#2})}
\newcommand{\thmref}[1]{Theorem~\ref{#1}}
\newcommand{\dfnref}[1]{Definition~\ref{#1}}
\newcommand{\exampleref}[1]{Example~\ref{#1}}

\excludecomment{hiddenproof}

\title{Selecting Computations: Theory and Applications} 
\author{ {\bf Nicholas Hay} and {\bf Stuart Russell} \\   
Computer Science Division \\  
University of California\\ 
Berkeley, CA 94720\\
\texttt{\{nickjhay,russell\}@cs.berkeley.edu}
\And 
{\bf David Tolpin} and {\bf Solomon Eyal Shimony} \\ 
Department of Computer Science \\ 
Ben-Gurion University of the Negev\\
Beer Sheva 84105, Israel\\
\texttt{\{tolpin,shimony\}@cs.bgu.ac.il}
}

\begin{document}

\maketitle 

\input{abstract}

\section{Introduction}

\input{introduction}

%% \input{related}

\section{On optimal policies for selection}\label{sec:optimal}

\input{optimalpolicies}

\section{Context effects and non-indexability}\label{sec:context}

\input{context}

\section{The blinkered policy}\label{sec:blinkered}\label{approx-bayesian-section}

\input{blinkered}

%% \section{Regret bounds and approximate policies}

%% [[Regret models: simple regret, regret with cost per sampling; regret goes to zero as c does]]

%% [[Expected simple regret bounds for normal case?]]

%% [[Blinkered sampling]]

%% [[ESPb: Frazier's continuous time approximation]]

\section{Upper bounds on Value of Information}\label{approx-nonbayesian-section}

%% Solomon: Text presumably should be placed after the relevant theorems.

\begin{comment}
	Theorem \ref{thm:optimal-myopic} provides necessary conditions
	for a stopping condition for the optimal policy in terms of a myopic
	policy, but that does not preclude premature stopping of a greedy policy.
	Conversely, Theorem \ref{thm:myopic-optimal} provides sufficient conditions, but these 
	conditions are hard to evaluate in practice.

	The myopic selection policy uses
	the intrinsic value of information (VOI) $\Lambda_i$ of testing an arm~$i$, which is
	the expected reward of the best arm given the information obtained by the test,
	minus the expected reward of the best arm according to the current information state.
	Unfortunately, the pure \textit{myopic} VOI estimate is of little use in
	Monte-Carlo sampling, since the effect of a single sample is small,
	and the myopic VOI estimate will often be zero.

	Instead, we generalize myopic policies to include ``semi-myopic"
	up to $N$ samples of the same arm; these are called ``blinkered" policies.
	Such policies allow potentially unlimited lookahead, but only in a single ``direction'' (one specific arm),
	as if we ``had our blinkers on''.	
\end{comment}

%% Solomon: end text to be (possibly moved).

In many practical applications of the selection problem, such as search in
the game of Go, prior distributions are unavailable.\footnote{The analysis is also applicable to
some Bayesian settings, using ``fake" samples to simulate prior distributions.}
In such cases, one can still bound
the value of information of myopic policies using {\em concentration
inequalities} to derive distribution-independent bounds on the
VOI. We obtain such bounds under the
following assumptions:
\begin{enumerate}
\item Samples are iid given the value of the arms (variables), as in the Bayesian schemes such as Bernoulli
sampling.
\item The expectation of a selection in a belief state is equal to the sample mean (and therefore,
   after sampling terminates, the arm with the greatest sample mean will be selected).
\end{enumerate}

When considering possible samples in the blinkered semi-myopic setting,
two cases are possible: either
	the arm $\alpha$ with the highest sample mean $\overline
  	X_\alpha$ is tested, and $\overline X_\alpha$ becomes lower than
 	$\overline X_\beta$ of the second-best arm $\beta$;
or, 
	another arm~$i$ is tested, and $\overline X_i$ becomes higher
    than $\overline X_\alpha$.

Our bounds below are applicable to any bounded distribution (without loss of generality 
bounded in $[0,1]$). Similar
bounds can be derived for certain unbounded distributions, such as the
normally distributed prior value with normally distributed
sampling.
We derive a VOI bound for testing an arm a fixed $N$ times,
where $N$ can be the remaining budget of available samples or
any other integer quantity.
Denote by  $\Lambda_i^b$ the intrinsic VOI of testing the $i$th arm
$N$ times, and the number of
samples already taken from the $i$th arm by $n_i$.
\begin{thm} $\Lambda_i^b$ is bounded from above as
\begin{align}
\label{eqn:thm-be}
  \Lambda_\alpha^b&\le \frac {N \overline X_\beta^{n_\beta}} {n_\alpha} \Pr(\overline X_\alpha^{n_\alpha+N}\le\overline X_\beta^{n_\beta})\nonumber\\
\Lambda_{i|i\ne\alpha}^b&\le \frac {N(1-\overline X_\alpha^{n_\alpha})} {n_i}\Pr(\overline   X_i^{{n_i}+N}\ge\overline X_\alpha^{n_\alpha})
\end{align}
\vspace{-24pt}
\label{thm:be}
\end{thm}
\begin{hiddenproof}
	\vspace{-2em}
	\begin{proof} For the case $i\ne \alpha$, the probability that the
	  $i$th arm is finally chosen instead of $\alpha$ is
	  $\Pr(\overline X_i^{n_i+N} \ge \overline X_\alpha^{n_\alpha})$. $X_i \le 1$,
	  therefore $\overline X_i^{n_i+N}\le \overline
	  X_\alpha^{n_\alpha}+\frac {N(1-\overline X_\alpha^{n_\alpha})} {N+n_i}$. Hence, the intrinsic value of blinkered
	  information is at most: 
	\begin{align}
	\label{eq:simplistic}
	\frac{ N(1-\overline  X_\alpha^{n_\alpha})}
	  {N+n_i}&\Pr(\overline X_i^{{n_i}+N}\hspace{-0.5em}\ge\overline X_\alpha^{n_\alpha})\nonumber \\
	&\le\frac{ N(1-\overline  X_\alpha^{n_\alpha})}
	{n_i}\Pr(\overline X_i^{{n_i}+N}\hspace{-0.5em}\ge\overline X_\alpha^{n_\alpha})
	\end{align}
	  Proof for the case $i=\alpha$ is similar.
	\end{proof}		
\end{hiddenproof}

The probabilities can be bounded from above using the
Hoeffding inequality \citep{Hoeffding.ineq}:
\begin{thm} The probabilities in \eqref{eqn:thm-be} are bounded from above as
\begin{align}
  \label{eqn:probound-blnk-hoeffding}
  \Pr&(\overline X_\alpha^{{n_\alpha}+N}\hspace{-6pt} \le \overline X_\beta^{n_\beta})
  \le 2\exp\left(- \varphi (\overline X_\alpha^{n_\alpha} - \overline X_\beta^{n_\beta})^2 n_\alpha
  \right)\nonumber\\
  \Pr&(\overline X_{i|i\ne\alpha}^{n_\alpha+N}\hspace{-6pt} \ge \overline X_\beta^{n_\beta})
  \le 2\exp\left(- \varphi (\overline X_\alpha^{n_\alpha} -\overline  X_i^{n_i})^2 n_i \right)\hspace{-6pt}
\end{align}
where $\varphi=\min \left(2(\frac {1+n/N} {1+\sqrt {n/N}})^2\right)=8(\sqrt 2 - 1)^2 > 1.37$.
\label{thm:hoeffding-prob-bounds}
\end{thm}

\begin{hiddenproof}
	\begin{proof}
	\eqref{eqn:probound-blnk-hoeffding}) follow from the
	observation that if $i\ne\alpha$, $\overline X_i^{n_i+N}>\overline X_\alpha^{n_i}$
	if and only if the mean $\overline X_i^N$ of $N$ samples from $n_i+1$
	to $n_i+N$ is at least $\overline X_\alpha^{n_i}+(\overline X_\alpha^{n_i}-\overline
	X_i^{n_i})\frac {n_i} N$.

	For any $\delta$, the probability that $\overline X_i^{n_i+N}$ is greater
	than $\overline X_\alpha^{n_i}$ is less than the probability that
	$\IE[X_i]\ge\overline X_i^n+\delta$
	\emph{or} $\overline X_i^N\ge \IE[X_i]+\overline X_\alpha^{n_\alpha}
	- \overline X_i^{n_i} - \delta +(\overline X_\alpha^{n_\alpha} - \overline X_i^{n_i})\frac {n_i} N$,
	thus, by the union bound, less than the sum of the probabilities:
	\begin{align}
	\Pr&(\overline X_i^{n_i+N}\ge \overline X_\alpha^{n_i})\nonumber\\
	   &\le\Pr(\IE[X_i]-\overline X_i^{n_i} \ge \delta)\\
	   &\hspace{0.25em}+\Pr\left(\overline X_i^N\hspace{-0.5em} - \IE[X_i] \ge \overline X_\alpha^{n_\alpha}\hspace{-0.5em}
	           - \overline X_i^{n_i}\hspace{-0.5em} - \delta +(\overline X_\alpha^{n_\alpha}\hspace{-0.5em} - \overline X_i^{n_i})\frac {n_i} N\right)\nonumber
	\end{align}
	Bounding the probabilities on the right-hand side using the Hoeffding
	inequality, obtain:
	\begin{align}
	\Pr&(\overline X_i^{n_i+N}\ge \overline X_\alpha^{n_\alpha})\le \nonumber\\
	    &\exp(-2\delta^2n_i)+\nonumber\\
	    &\quad\exp\left(-2\left((\overline X_\alpha^{n_\alpha}\hspace{-0.5em}
	         - \overline X_i^{n_i})\left(1+\frac {n_i} N\right)
	         - \delta\right)^2N\right)
	\label{eqn:app-hoeffding-le-maxexp}
	\end{align}
	Find $\delta$ for which the two terms on the right-hand side of
	\eqref{eqn:app-hoeffding-le-maxexp} are equal:
	\begin{equation}
	\exp(-\delta^2n) = \exp\left(-2\left((\overline X_\alpha - \overline X_i)(1+\frac {n_i} N) - \delta\right)^2N\right)\label{eqn:app-hoeffding-eq-exp}
	\end{equation}
	Solve \eqref{eqn:app-hoeffding-eq-exp} for $\delta$: $\delta=\frac {1+\frac {n_i} N} {1+\sqrt {\frac {n_i} N}} (\overline X_\alpha^{n_\alpha}\hspace{-0.5em}
	- \overline X_i^{n_i}) \ge 2(\sqrt 2 - 1)(\overline X_\alpha^{n_\alpha}\hspace{-0.5em}-\overline X_i^{n_i})$. Substitute $\delta$ into 
	\eqref{eqn:app-hoeffding-le-maxexp} and obtain
	\begin{align}
	\Pr&(\overline X_i^{n_i}\ge \overline X_\alpha^{n_\alpha}) \nonumber\\
	& \le 2\exp\left(-2\left( \frac {1+\frac {n_i} N} {1+\sqrt {\frac {n_i} N}}
	                          (\overline X_\alpha^{n_\alpha} - \overline X_i^{n_i})\right)^2 n_i\right)\nonumber \\
	& \le 2\exp(-8(\sqrt 2 - 1)^2(\overline X_\alpha^{n_\alpha} - \overline X_i^{n_i})^2n_i)\nonumber\\
	& = 2\exp(-\varphi(\overline X_\alpha^{n_\alpha} - \overline X_i^{n_i})^2n_i)
	\end{align}
	Derivation for the case $i=\alpha$ is similar.
	\end{proof}	
\end{hiddenproof}

\begin{crl}
An upper bound on the VOI estimate $\Lambda_i^b$ is obtained
by substituting \eqref{eqn:probound-blnk-hoeffding} into (\ref{eqn:thm-be}).
\begin{align}
  \label{eqn:bound-blnk-hoeffding}
  \Lambda&_\alpha^b \le \hat\Lambda_\alpha^b=\frac {2N\overline X_\beta^{n_\beta}} {n_\alpha}\exp\left(- \varphi(\overline X_\alpha^{n_\alpha}\hspace{-0.5em} - \overline X_\beta^{n_\beta})^2 n_\alpha\right)\\
  \Lambda&_{i|i\ne\alpha}^b\le \hat\Lambda_i^b=  \frac {2N(1-\overline  X_\alpha^{n_\alpha})} {n_i}\exp\left(- \varphi(\overline X_\alpha^{n_\alpha}\hspace{-0.5em} - \overline X_i^{n_i})^2 n_i\right)\nonumber
\end{align}
\label{crl:bound-blnk-hoeffding}
\end{crl}
\vspace{-2em}

More refined bounds can be obtained through tighter estimates on the
probabilities in \eqref{eqn:thm-be}, for example, based on the empirical Bernstein
inequality~\citep{MaurerPontil.benrstein}, or through a more careful
application of the Hoeffding inequality, resulting in:
\begin{align}
\Lambda_i^b&\le\frac {N\sqrt \pi} {n_i \sqrt {n_i}}
  \left[\mathrm{erf}\left((1\hspace{-0.25em}-\hspace{-0.25em}\overline X_i^{n_i})\sqrt {n_i}\right)
      \hspace{-0.25em}-\hspace{-0.25em}\mathrm{erf}\left((\overline X_\alpha^{n_\alpha}\hspace{-0.5em} - \overline X_i^{n_i})\sqrt{n_i}\right)\right]\nonumber\\
\Lambda_\alpha^b&\le\frac {N\sqrt \pi} {n_\alpha \sqrt {n_\alpha}}
  \left[\mathrm{erf}\left(\overline X_\alpha^{n_\alpha}\sqrt {n_\alpha}\right)
      -\mathrm{erf}\left((\overline X_\alpha^{n_\alpha}\hspace{-0.5em} - \overline X_\beta^{n_\beta})\sqrt{n_\alpha}\right)\right]
\label{eqn:erf-blinkered}
\end{align}

Selection problems usually separate out the decision of {\em whether} to
sample or to stop (called the stopping policy), and {\em what} to sample.
We'll examine the first issue here, along with the empirical evaluation 
of the above approximate algorithms, and the second in the following section.

Assuming that the sample costs are constant,
a semi-myopic policy will decide to test the arm that has the best
current VOI estimate. 
When the distributions are unknown, it makes sense
to use the upper bounds established in \thmref{thm:be}, as we do in the following.
This evaluation assumes a fixed budget of samples, which is
completely used up by each of the candidate schemes, making a stopping
criterion irrelevant.

\begin{figure}[h!]
\centering
\includegraphics[scale=0.55]{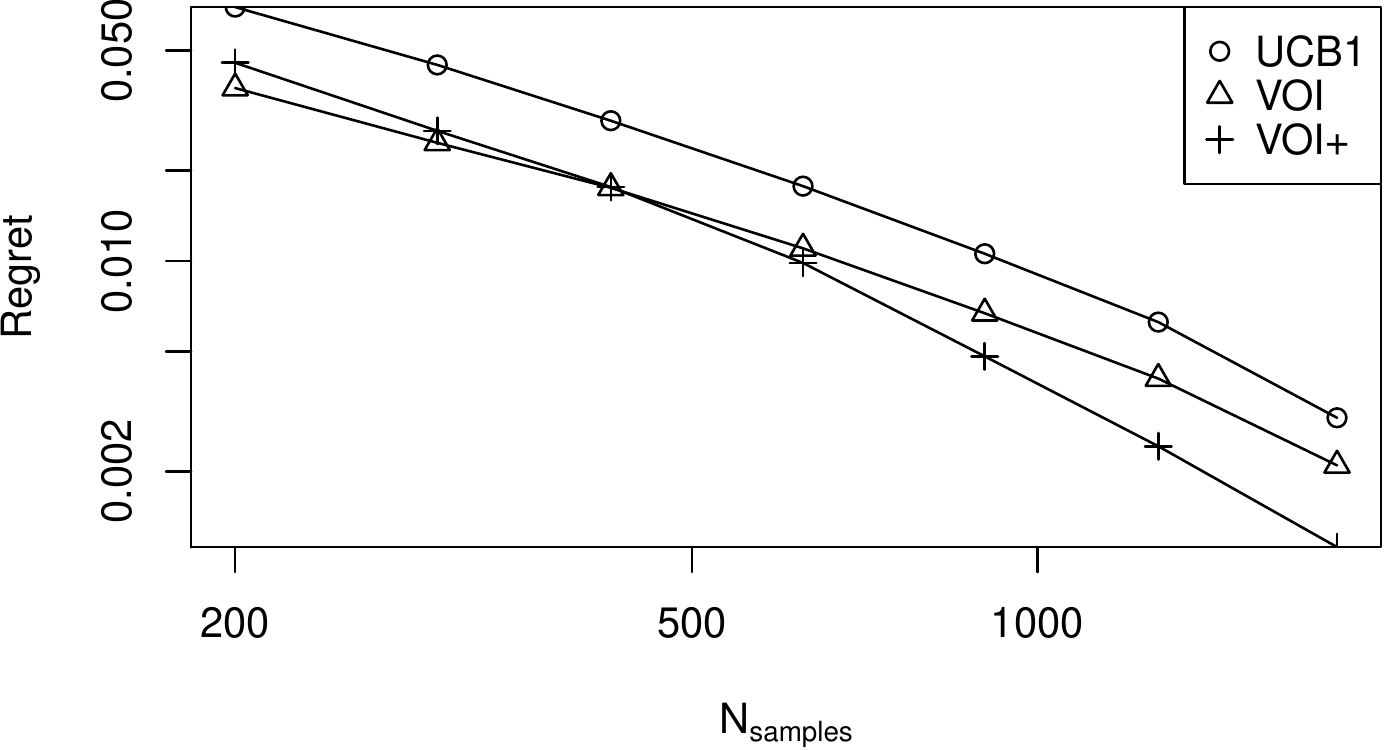}
\caption{Average regret of various policies as a function of the fixed number 
of samples in a 25-action Bernoulli sampling problem, over 10000 trials.}
\label{fig:random-instances}
\end{figure}

The sampling policies are compared on random Bernoulli
selection problem instances. \figref{fig:random-instances} shows results for
randomly-generated selection problems with 25 Bernoulli arms, where
the mean rewards of the arms are distributed uniformly in~$[0,1]$, 
for a range of sample budgets~$200..2000$, with multiplicative
step of~$2$, averaging over 10000 trials.  We compare UCB1 with the 
policies based on the bounds in
\eqref{eqn:bound-blnk-hoeffding} (VOI) and
\eqref{eqn:erf-blinkered} (VOI+).
UCB1 is always considerably worse than the VOI-aware sampling policies.

\section{Sampling in trees}
\label{sec:empirical-evaluation}\label{mcts-section}

The previous section addressed the selection problem in the flat case.
Selection in trees is more complicated.  The goal of Monte-Carlo tree 
search \citep{Chaslot.montecarlo} at the root node 
is usually to select an action that appears to be the best based on outcomes
of \textit{search rollouts}.
But the goal of rollouts at non-root nodes
is different than at the root: here it is important to better approximate the
value of the node, so that selection at the root can be more informed. The exact analysis
of sampling at internal nodes is outside the scope of this paper. At present we 
have no better proposal for internal nodes than to use UCT there.

We thus propose the following hybrid sampling scheme \citep{TolpinShimony.mcts}: 
	at the \emph{root node}, sample based on the VOI estimate;
	at \emph{non-root nodes}, sample using UCT.

Strictly speaking, even at the root node the stationarity assumptions\footnote{This is not a restriction,
however, of the general formalism in \secref{sec:optimal}.} 
underlying our belief-state
MDP for selection do not hold exactly. UCT is an adaptive scheme, and therefore the values
generated by sampling at non-root nodes will typically cause values observed at
children of the root node to be non-stationary. 
Nevertheless, sampling based on VOI estimates
computed as for stationary distributions works well in
practice. As illustrated
by the empirical evaluation (\secref{sec:empirical-evaluation}),
estimates based on upper bounds on the VOI result in good sampling
policies, which exhibit performance comparable to the performance of
some state-of-the-art heuristic algorithms.

\subsection{Stopping criterion}
\label{sec:control-stopping-criterion}

When a sample has a known cost commensurable with the value of
information of a measurement, an upper bound on the intrinsic VOI can also
be used to stop the sampling if the intrinsic VOI of any arm
is less than the total cost of sampling $C$: $\max_i \Lambda_i \le C$.

The VOI estimates of \eqrefs{eqn:thm-be}{eqn:bound-blnk-hoeffding} 
include the remaining sample budget $N$ as a
factor, but given the cost of a single sample $c$, the cost of the
remaining samples accounted for in estimating the intrinsic VOI is
$C=cN$. $N$ can be dropped on both sides of the inequality,
giving a reasonable stopping criterion:
\begin{align}
\frac 1 N \Lambda_\alpha^b \le&\frac {\overline X_\beta^{n_\beta}}
  {n_\alpha}\Pr(\overline X_\alpha^{n_\alpha+N}\le\overline
  X_\beta^{n_\alpha})\le c\nonumber\\
\frac 1 N \max_i\Lambda_i^b\le &\max_i\frac {(1-\overline X_\alpha^{n_\alpha})} {n_i}\Pr(\overline
  X_i^{n_i+N}\ge\overline X_\alpha^{n_\alpha})\le c\nonumber\\
    &\forall i: i\ne\alpha
\label{eqn:stopping-blnk}
\end{align}
The empirical evaluation (\secref{sec:empirical-evaluation})
confirms the viability of this stopping criterion and illustrates the
influence of the sample cost $c$ on the performance of
the sampling policy. When the sample cost $c$ is unknown, one can perform initial calibration experiments
to determine a reasonable value, as done in the following.

\subsection{Sample redistribution in trees}
\label{sec:control-redistribution}

The above hybrid approach assumes
that the information obtained from rollouts in the
current state is discarded after an real-world action is selected. In practice,
many successful Monte-Carlo tree search algorithms reuse rollouts
generated at earlier search states, if the sample traverses the
current search state during the rollout; thus, the value of information of a rollout is
determined not just by the influence on the choice of the action at
the current state, but also by its potential influence on the choice at future
search states.

One way to account for this reuse would be to incorporate the
`future' value of information into a VOI estimate. However, this 
approach requires a nontrivial extension of the theory of metareasoning for search.
Alternately, one can behave myopically with respect to the search tree depth:
\begin{enumerate}
\item Estimate VOI as though the information is discarded after each step,
\item Stop early if the VOI is below a certain threshold
   (see \secref{sec:control-stopping-criterion}), and
\item Save the unused sample budget for search in future states, such that
   if the nominal budget is $N$, and the unused budget in the last state
   is $N_u$, the search budget in the next state will be $N+N_u$.
\end{enumerate}
In this approach, the cost $c$ of a sample in the current state is the
VOI of increasing the budget of a future state by one sample.  It is
unclear whether this cost can be accurately estimated, but supposing
a fixed value for a given problem type and algorithm implementation
would work. Indeed, the empirical evaluation (\secref{sec:emp-go})
confirms that stopping and sample redistribution based on a learned
fixed cost  substantially improve the performance of the VOI-based
sampling policy in game tree search.

\subsection{Playing Go against UCT}
\label{sec:emp-go}

The hybrid policies were compared on the game Go, a search domain
in which UCT-based MCTS has been particularly successful
\citep{Gelly.mogo}. A modified version of Pachi \citep{Braudis.pachi}, a state of the art
Go program, was used for the experiments:
\begin{itemize}
\item The UCT engine of Pachi was extended with VOI-aware sampling
  policies at the first step. 
\item The stopping criterion for the VOI-aware policy was
  modified and based solely on the sample cost, specified as
  a constant parameter. The heuristic stopping criterion for the
  original UCT policy was left unchanged.
\item The time-allocation model based on the fixed number of samples
  was modified for \textit{both the original UCT policy and the VOI-aware
  policies} such that 
  \begin{itemize}
    \item Initially, the same number of samples is available to
      the agent at each step, independently of the number of pre-simulated
      games;  
    \item If samples were unused at the current step,
      they become available at the next step. 
  \end{itemize}
\end{itemize}
While the UCT engine is not the most powerful engine of Pachi, it is still a strong
player. On the other hand, additional features of more advanced
engines would obstruct the MCTS phenomena which are the subject of
the experiment.
\begin{figure}[h!]
\centering
\includegraphics[scale=0.55]{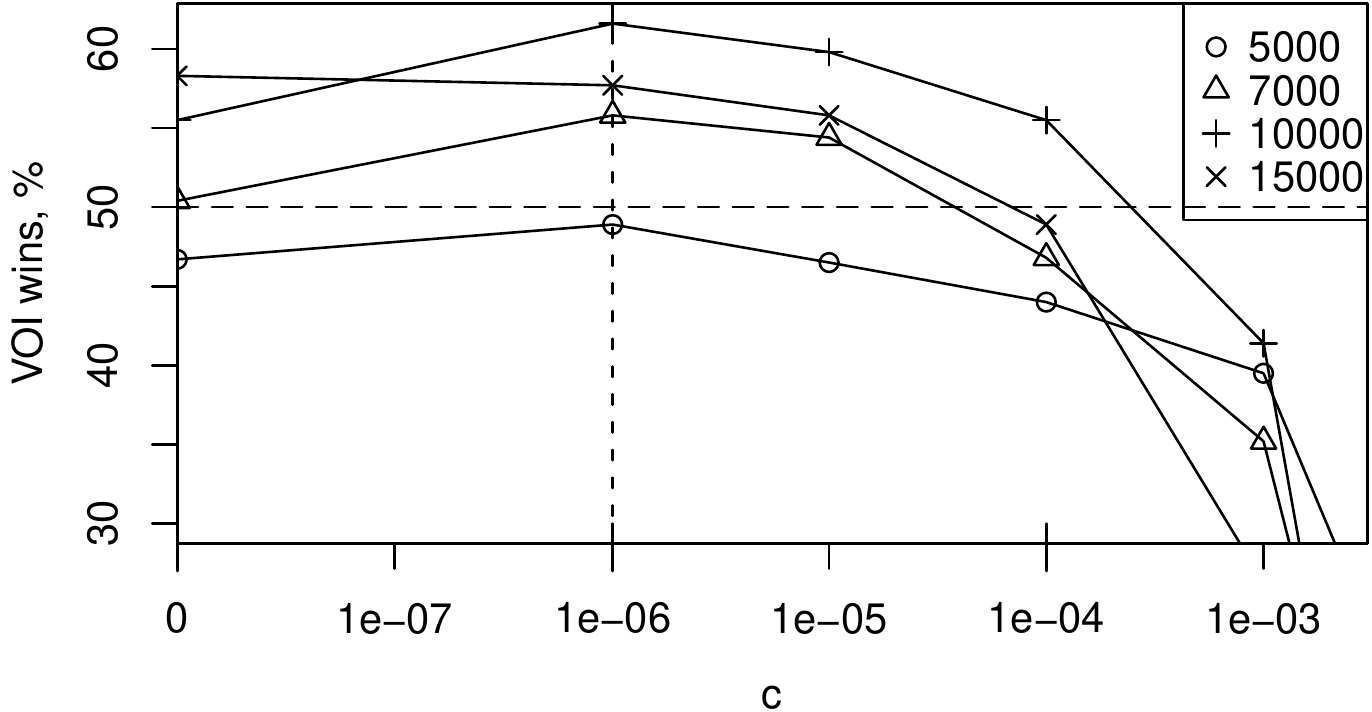}
\caption{Winning rate of the VOI-aware policy in Go as a function of the cost $c$, for varying numbers of samples per ply.}
\label{fig:uctvoi}
\end{figure}
The engines were compared on the 9x9 board, for 5000, 7000, 1000, and
15000 samples (game simulations) per ply, each experiment repeated
1000 times. \figref{fig:uctvoi} depicts a calibration experiment,
showing the winning rate of the VOI-aware policy against UCT as a function of
the stopping threshold $c$ (if the maximum VOI of a sample is below
the threshold, the simulation is stopped, and a move is chosen). Each
curve in the figure corresponds to a certain number of samples per
ply.  For the stopping threshold of $10^{-6}$, the VOI-aware policy
is almost always better than UCT, and reaches the winning rate of
64\% for 10000 samples per ply.

\begin{figure}[h!]
\centering
\includegraphics[scale=0.55]{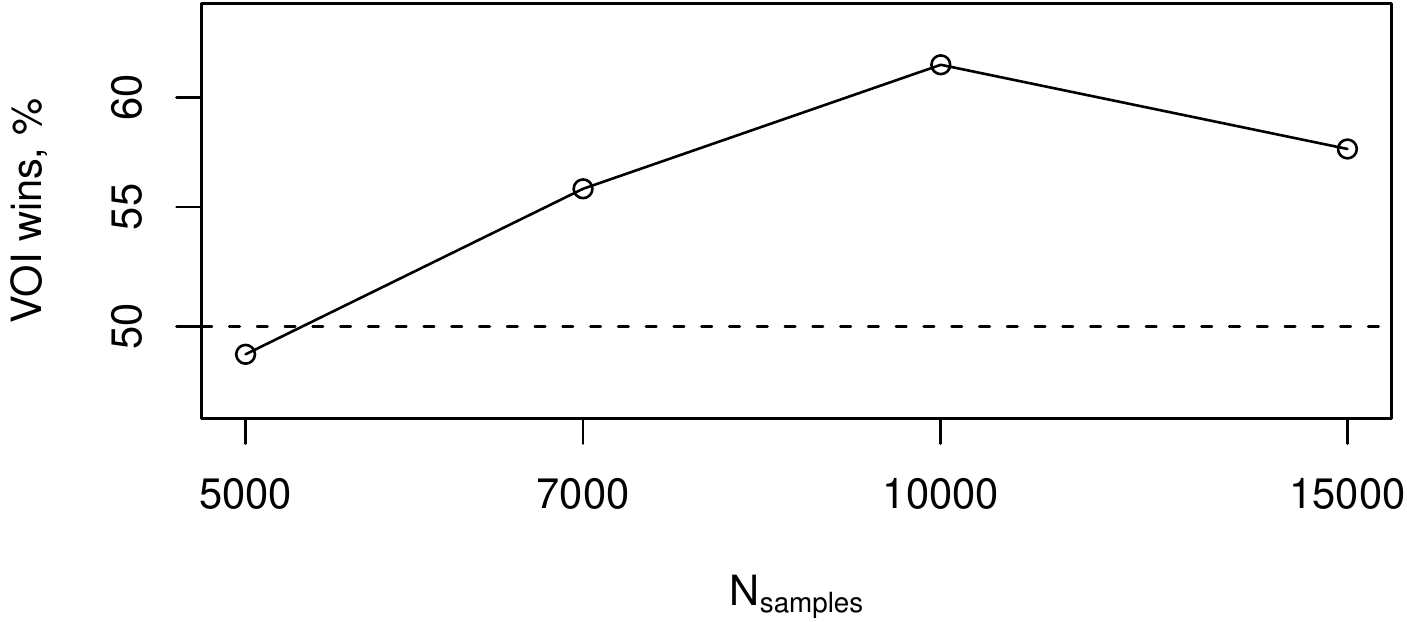}
\caption{Winning rate of the VOI-aware policy in Go as a function of the number of samples, fixing cost $c=10^{-6}$.}
\label{fig:voi-wins}
\end{figure}

\figref{fig:voi-wins}
shows the winning rate of VOI against UCT $c=10^{-6}$. In agreement with the intuition
(\figref{sec:control-redistribution}), VOI-based stopping and
sample redistribution is most influential for intermediate numbers of
samples per ply. When the maximum number of samples is too low, early
stopping would result in poorly selected moves. On the other hand,
when the maximum number of samples is sufficiently high, the VOI of
increasing the maximum number of samples in a future state is low.

Note that if we disallowed reuse of samples in both Pachi and
in our VOI-based scheme, the VOI based-scheme
win rate is even higher than shown in \figref{fig:voi-wins}. This is as expected,
as this setting (which is somewhat unfair to Pachi) is closer to
meeting the assumptions underlying the selection MDP.

\input{conclusion}

\input{acknowledgements}

\bibliographystyle{abbrvnat}
\bibliography{refs}

\end{document}

%% file: abstract.tex
\begin{abstract}
Sequential decision problems are often approximately solvable by
simulating possible future action sequences.  {\em Metalevel} decision
procedures have been developed for selecting {\em which} action
sequences to simulate, based on estimating the expected
improvement in decision quality that would result from any particular
simulation; an example is the recent work on using bandit algorithms
to control Monte Carlo tree search in the
game of Go.  In this paper we develop a theoretical basis for
metalevel decisions in the statistical framework of Bayesian {\em selection problems}, 
arguing (as others have done) that this is more appropriate than the bandit
framework.  We derive a number of basic results applicable to 
Monte Carlo selection problems, including the first finite sampling
bounds for optimal policies in certain cases; we also provide a simple
counterexample to the intuitive conjecture that an optimal policy will
necessarily reach a decision in all cases.  We then derive heuristic
approximations in both Bayesian and distribution-free settings and
demonstrate their superiority to bandit-based heuristics in one-shot
decision problems and in Go.
\end{abstract}

%% file: introduction.tex
The broad family of sequential decision problems includes
combinatorial search problems, game playing, robotic path planning,
model-predictive control problems, Markov decision processes (MDP), whether fully or
partially observable, and a huge range of applications. In almost all
realistic instances, exact solution is intractable and approximate
methods are sought. Perhaps the most popular approach is to simulate
a limited number of possible future action sequences, 
in order to find a move in the current state that is (hopefully)
near-optimal. In this paper, we develop a theoretical framework to examine the problem of selecting
{\em which} future sequences to simulate.
We derive a number of new results concerning optimal policies for this selection problem
as well as new heuristic policies for controlling Monte Carlo simulations.
As described below, these policies outperform previously published methods for
``flat'' selection and game-playing in Go.

The basic ideas behind our approach are best explained in a familiar
context such as game playing.  A typical game-playing algorithm
chooses a move by first exploring a tree or graph of move sequences
and then selecting the most promising move based on this exploration.
Classical algorithms typically explore in a fixed order, imposing a
limit on exploration depth and using pruning methods to avoid
irrelevant subtrees; they may also reuse some previous computations
(see Section \ref{sec:control-redistribution}).  Exploring unpromising
or highly predictable paths to great depth is often wasteful; for a
given amount of exploration, decision quality can be improved by
directing exploration towards those actions sequences whose outcomes
are helpful in selecting a good move. Thus, the {\em metalevel}
decision problem is to choose what future action sequences to explore
(or, more generally, what deliberative computations to do), while the
{\em object-level} decision problem is to choose an action to execute
in the real world.

That the metalevel decision problem can itself be formulated and
solved decision-theoretically was noted by \citet{Matheson:1968},
borrowing from the related concept of {\em information value
  theory}~\citep{Howard:1966}. In essence, computations can be
selected according to the expected improvement in decision quality
resulting from their execution. I. J.~\citet{Good:1968} independently
proposed using this idea to control search in chess, and later defined
``Type II rationality'' to refer to agents that optimally solve the
metalevel decision problem before acting. As interest in probabilistic
and decision-theoretic approaches in AI grew during the 1980s, several
authors explored these ideas further~\citep{Dean+Boddy:1988,Doyle:1988,Fehling+Breese:1988,Horvitz:1987b}.
Work by
\citet{Russell+Wefald:1988a,Russell+Wefald:1991a,Russell+Wefald:1991b}
formulated the metalevel sequential decision problem, employing an
explicit model of the results of computational actions, and applied
this to the control of game-playing search in Othello with encouraging
results.

An independent thread of research on metalevel control began with work
by \citet{Kocsis+Szepesvari:2006} on the UCT algorithm, which operates
in the context of {\em Monte Carlo tree search} (MCTS) algorithms.
In MCTS, each computation takes the form
of a simulation of a randomized sequence of actions leading from a leaf of the
current tree to a terminal state. UCT is primarily a method for
selecting a leaf from which to conduct the next simulation, and
forms the core of the successful \textsc{MoGo} algorithm for Go 
playing \citep{Gelly+Silver:2011}.  The UCT algorithm is
based on the the theory of bandit problems \citep{Berry+Fristedt:1985} and the asymptotically near-optimal
UCB1 bandit algorithm \citep{Auer+et+al:2002}. UCT applies
UCB1 recursively to select actions to perform within simulations.

It is natural to consider whether the two independent threads are
consistent; for example, are bandit algorithms such as UCB1
approximate solutions to some particular case of the metalevel
decision problem defined by Russell and Wefald? The answer, perhaps
surprisingly, is no.  The essential difference is that, in bandit
problems, every trial involves executing a real object-level action
with real costs, whereas in the metareasoning problem the trials are
{\em simulations} whose cost is usually independent of the utility of
the action being simulated.  Hence, as \citet{Audibert+al:2010}
and \citet{Bubeck+al:2011} have also noted, UCT applies bandit
algorithms to problems that are not bandit problems.

One consequence of the mismatch is that bandit policies are
inappropriately biased away from exploring actions whose current
utility estimates are low.  Another consequence is the absence of any
notion of ``stopping'' in bandit algorithms, which are designed for
infinite sequences of trials.  A metalevel policy needs to decide
when to stop deliberating and execute a real action.

Analyzing the metalevel problem within an appropriate theoretical
framework ought to lead to more effective algorithms than those
obtained within the bandit framework.  For Monte Carlo computations,
in which samples are gathered to estimate the utilities of actions,
the metalevel decision problem is an instance of the {\em selection
problem} studied in statistics~\citep{Bechhofer:1954,Swisher+et+al:2003}.  Despite
some recent work \citep{Frazier+Powell:2010,TolpinShimony:2012}, the theory of selection
problems is less well understood than that of bandit problems.
Most work has focused on the probability of selection error rather than
optimal policies in the Bayesian setting~\citep{Bubeck+al:2011}.
Accordingly, we present in \secrefs{sec:optimal}{sec:context}
a number of results concerning optimal policies for the general case
as well as specific finite
bounds on the number of samples collected by optimal policies for
Bernoulli arms with beta priors.  We also provide a simple
counterexample to the intuitive conjecture that an optimal policy
should not spend more on deciding than the decision is worth; in fact,
it is possible for an optimal policy to compute forever. We also show
by counterexample that optimal {\em index
policies}~\citep{Gittins:1989} may not exist for selection
problems.

Motivated by this theoretical analysis, we propose in \secrefs{approx-bayesian-section}{approx-nonbayesian-section}
two families of heuristic approximations, one for the
Bayesian case and one for the distribution-free setting.
We show empirically that these rules give better performance than UCB1 on 
a wide range of standard (non-sequential) selection problems.
\secref{mcts-section} shows similar results for the case of guiding Monte Carlo tree search
in the game of Go.

%% (Proofs omitted due to space constraints can be found in the supplementary material.)

%% file: optimalpolicies.tex
%% \section{On optimal policies for selection}

%% Define selection problem and metalevel MDP representation.

%% \note{Unsure about terminology here: selection problems, sampling problems, 
%% metalevel probability model, metalevel decision problem (conflicts with Markov decision process?).}

%% ============================
%% \subsection{Selection problems}
%% ============================

%% \note{Formal definition of selection problems and the metalevel MDP with cost per sample (time value); also mention budgeted learning.}

In a selection problem the decision maker is faced with a choice among
alternative arms\footnote{Alternative actions are known as \emph{arms} in the bandit setting;
we borrow this terminology for uniformity.}.  To make this choice, they may gather evidence about the
utility of each of these alternatives, at some cost.  The objective is to maximize
the {\em net utility}, i.e., the expected utility of the final arm selected, less the cost of gathering the evidence. 
In the classical case~\citep{Bechhofer:1954}, evidence might consist of physical samples
from a product batch; in a metalevel problem with Monte Carlo simulations,
the evidence consists of outcomes of sampling computations:

\begin{dfn} \label{dfn:metalevel-model}
	A \term{metalevel probability model} is a tuple $(U_1,\dots,U_k,\Evidence)$ 
	consisting of jointly distributed random variables:
	\begin{itemize}
		\item Real random variables $U_1,\dots,U_k$, where $U_i$ is the utility of arm~$i$, and
		\item A countable set $\Evidence$ of random variables, each variable $E\in\Evidence$ being 
		      a computation that can be performed and whose value is the result of that computation.
	\end{itemize}
\end{dfn}
%% Not actually true that we necessarily define this with U's as roots:  in sampling models they are naturally /leaves/.
%% Bounedness assumption
For simplicity, in the below we'll assume the utilities $U_i$ are bounded, 
without loss of generality in $[0,1]$.  We will abuse notation and denote by $e\in E$ that $e$ is
a potential value of the computation $E$.

\begin{example}[Bernoulli sampling]\label{example:bernoulli}
%% One simple metalevel probability model is used as a model of the results of simulations \citep{someone}.
In the \term{Bernoulli metalevel probability model},
each arm will either succeed or not $U_i\in\{0,1\}$, with an unknown latent frequency of success $\Theta_i$, 
and a set of stochastic simulations of possible consequences
$\mathcal{E} = \{E_{ij} | 1\le i \le k, j\in \mathbb{N}\}$ that can be performed:
\begin{align*}
	\Theta_i &\iid {\rm Uniform}[0,1]                      &&\quad\text{for $i\in\{1,\dots,k\}$}\\
	U_i \given \Theta_i &\sample {\rm Bernoulli}(\Theta_i) &&\quad\text{for $i\in\{1,\dots,k\}$}\\
	E_{ij} \given \Theta_i &\iid {\rm Bernoulli}(\Theta_i) &&\quad\text{for $i\in\{1,\dots,k\}$, $j\in\Nat$}
\end{align*}

The \term{one-armed Bernoulli metalevel probability model} has $k=2$,
$\Theta_1=\lambda\in[0,1]$ a constant, and $\Theta_2\sample{\rm Uniform}[0,1]$.
\end{example}

A metalevel probability model, when combined with a cost of computation $c>0$,%
\footnote{The assumption of a fixed cost of computation is a simplification; 
	precise conditions for its validity are given by \citet{Harada:1997}.} 
defines a metalevel decision problem: what is the optimal strategy with which to choose a sequence 
of computations $E\in\Evidence$ in order to maximize the agent's net utility?
Intuitively, this strategy should choose the computations that give the most evidence relevant
to deciding which arm to use, stopping when the cost of computation 
outweighs the benefit gained. We formalize the selection problem as a Markov Decision Process
(see, e.g., \citet{Puterman:1994}):

\begin{dfn}
A (countable state, undiscounted) \term{Markov Decision Process} (MDP) is a tuple $M=(S,s_0,A_s,T,R)$ where:
	$S$ is a countable set of states,
	$s_0\in S$ is the fixed initial state,
	$A_s$ is a countable set of actions available in state $s\in S$,
	$T(s,a,s')$ is the transition probability from $s\in S$ to $s'\in S$ after performing action $a\in A_s$,
	and $R(s,a,s')$ is the expected reward received on such a transition.
\end{dfn}

To formulate the metalevel decision problem as an MDP, we define the states as sequences of
computation outcomes and allow for a terminal state when the agent chooses to stop computing and act:

\begin{dfn}\label{dfn:metalevel-mdp}
	Given a metalevel probability model%
		\footnote{Definition~\ref{dfn:metalevel-model} made no assumption about the computational result
			variables $E_i\in\Evidence$, but for simplicity in the following we'll assume that
			each $E_i$ takes one of a countable set of values.  Without loss of generality, 
			we'll further assume the domains of the computational variables $E\in\Evidence$ are disjoint.}
	$(U_1,\dots,U_k,\Evidence)$ and
	a cost of computation $c>0$, a corresponding \term{metalevel decision problem}
	is any MDP $M=(S,s_0,A_s,T,R)$ such that
	\begin{align*}
		S &= \{\bot\}\cup\{\langle e_1\dots, e_n \rangle : e_i\in E_i \text{ for all $i$,} \\
								& \qquad\qquad \text{for finite $n\ge0$ and distinct $E_i\in\Evidence$}\} \\
		s_0 &= \langle \rangle \\
		A_s &= \{\bot\}\cup\Evidence_s \\
	\intertext{where $\bot\in S$ is the unique terminal state,
	where $\Evidence_s\subseteq\Evidence$ is a state-dependent subset of allowed computations,
	and when given any $s=\langle e_1, \dots, e_n \rangle\in S$,
	computational action $E\in\Evidence$, 
	and $s'= \langle e_1, \dots, e_n, e \rangle\in S$ where $e\in E$, we have:}
		T(s,E,s') &= P(E=e \given E_1=e_1,\dots,E_n=e_n) \\
		T(s,\bot,\bot) &= 1 \\
		R(s,E,s') &= -c \\
		R(s,\bot,\bot) &= \max_i \mu_i(s) % \max_i \IE[U_i \given E_1=e_1,\dots,E_n=e_n]
	\end{align*}		
	where $\mu_i(s) = \IE[U_i \given E_1=e_1,\dots,E_n=e_n]$.
\end{dfn}

Note that when stopping in state $s$, the expected utility of 
action $i$ is by definition $\mu_i(s)$, so the optimal action to take is $i^*\in\argmax_i \mu_i(s)$
which has expected utility $\mu_{i*}(s) = \max_i\mu_i(s)$.

One can optionally add an external constraint on the number of computational actions, or their total cost,
in the form of a deadline or {\em budget}.  This bridges with the related area of budgeted learning \citep{Madani+et+al:2004}.
Although this feature is not formalized in the MDP, it can be added by including either time or past total cost 
as part of the state.

\begin{example}[Bernoulli sampling]\label{example:bernoulli2}
In the Bernoulli metalevel probability model (\exampleref{example:bernoulli}),
note that: 
\begin{align}
	\Theta_i \given E_{i1},\dots,E_{in_i} &\sim {\rm Beta}(s_i+1, f_i+1)  \label{eq:bernoulli1}\\
	E_{i(n_i+1)} \given E_{i1},\dots,E_{in_i} &\sim {\rm Bernoulli}\left(\frac{s_i+1}{n_i+2}\right) \label{eq:bernoulli2} \\
	\IE[U_i \given E_{i1},\dots,E_{in_i}] &= (s_i+1)/(n_i+2) \label{eq:bernoulli3}
\end{align}
by standard properties of these distributions, where $s_i=\sum_{j=1}^{n_i} E_{in_i}$
is the number of simulated successes of arm~$i$, and $f_i=n_i-s_i$ the failures.  By \eqref{eq:bernoulli1}, 
the state space is the set of all $k$ pairs $(s_i,f_i)$; \eqrefs{eq:bernoulli2}{eq:bernoulli3}
suffice to give the transition probabilities and terminal rewards, respectively.
The one-armed Bernoulli case is similar, requiring as state just $(s,f)$ defining the posterior over $\Theta_2$.
\end{example}

Given a metalevel decision problem $M=(S,s_0,A_s,T,R)$ one defines policies and 
value functions as in any MDP.  A (deterministic, stationary) \term{metalevel policy} 
$\pi$ is a function mapping states $s\in S$ to actions to take in that state $\pi(s)\in A_s$.	

The \term{value function} for a policy $\pi$ gives the expected total reward
received under that policy starting from a given state $s\in S$, and the \term{Q-function}
does the same when starting in a state $s\in S$ and taking a given action $a\in A_s$: 
\begin{equation}
		V^\pi_M(s) = \IE^\pi_M\left[ \sum_{i=0}^N R(S_i,\pi(S_i),S_{i+1}) \given S_0 = s\right]	\label{eq:value-fn}\\
	%%	Q^\pi_M(s,a) &= \IE^\pi_M\left[ \sum_{i=0}^N R(S_i,\pi(S_i),S_{i+1}) \given S_0 = s, A_0 = a\right] \label{eq:qvalue-fn}
\end{equation}
where $N\in[0,\infty]$ is the random time the MDP is terminated, i.e.,
the unique time where $\pi(S_N)=\bot$,
%% \footnote{Whenever $N<\infty$, $S_{N+1}=\bot$ the unique terminal state.
%% Instead of having a random termination time, one can equivalently make
%% the state $\bot$ absorbing with zero reward on all transitions.}
and similarly for the Q-function $Q^\pi_M(s,a)$.

As usual, an \term{optimal policy} $\pi^*$, when it exists, is one that maximizes 
the value from every state $s\in S$, i.e., if we define for each $s\in S$
\[
	V^*_M(s)   = \sup_\pi V^\pi_M(s),
%%	Q^*_M(s,a) &= \sup_\pi Q^*_M(s,a),
\]
then an optimal policy $\pi^*$ satisfies $V^{\pi^*}_M(s) = V^*_M(s)$
for all $s\in S$, where we break ties in favor of stopping.
%% and by standard results, $\pi^*(s) \in \argmax_{a\in A_s} Q^*_M(s,a)$, where 

The optimal policy must balance the cost of computations with the improved decision
quality that results.  This tradeoff is made clear in the value function:

\begin{thm}\label{thm:value-of-computation}
	The value function of a metalevel decision process $M=(S,s_0,A_s,T,R)$ is of the form
	\[
		V^\pi_M(s) = \IE^\pi_M[ -c\,N + \max_i \mu_i(S_N) \given S_0=s]
	\]
	where $N$ denotes the (random) total number of computations performed;
	similarly for $Q^\pi_M(s,a)$.
\end{thm}

\begin{hiddenproof}
	Follows immediately from \eqref{eq:value-fn} and the definition of the
	reward function in \dfnref{dfn:metalevel-mdp}.
\end{hiddenproof}

In many problems, including the Bernoulli sampling model of \exampleref{example:bernoulli2},
the state space is infinite. Does this preclude solving for the optimal policy?  Can 
infinitely many computations be performed? 

%%This immediately raises the question of whether
%%one can solve for an optimal policy, and whether it is sometimes optima
%%to perform infinitely many computations.

There is in full generality an upper bound on the \emph{expected} number of computations
a policy performs:

\begin{thm}\label{thm:bounded-expected-computations}
	The optimal policy's expected number of computations is bounded by the 
	value of perfect information \citep{Howard:1966} times the inverse cost $1/c$:
	\[
		\IE^{\pi^*}[N\given S_0=s] \le \frac{1}{c} \left(\IE[\max_i U_i\given S_0=s] - \max_i \mu_i(s)\right).
	\]
	Further, any policy $\pi$ with infinite expected number of computations 
	%% $\IE^\pi[N]=\infty$
	has negative infinite value, hence the optimal
	policy stops with probability one.
\end{thm}

\begin{hiddenproof}
	The first follows as in state $s$ the optimal policy has value at least that
	of stopping immediately ($\max_i \mu_i(s)$), and as $\IE \max_i\mu_i(S_N) \le \IE \max_i U_i$ by Jensen's inequality.
	The second from \thmref{thm:value-of-computation}.
\end{hiddenproof}

Although the \emph{expected} number of computations is always bounded,
there are important cases in which the \emph{actual} number is not, such as
the following inspired by the sequential probability ratio test \citep{Wald+1945}:

\begin{example}\label{example:sprt}
Consider the Bernoulli sampling model for two arms but with a different prior:
	$\Theta_1=1/2$,
	and $\Theta_2$ is $1/3$ or $2/3$ with equal probability.
Simulating arm~1 gains nothing, and after $(s,f)$ simulated successes and failures of arm~2
the posterior odds ratio is
\[
	\frac{P(\Theta_2=2/3\given s,f)}{P(\Theta_2=1/3\given s,f)} = \frac{(2/3)^s(1/3)^f}{(1/3)^s(2/3)^f}= 2^{s-f}.
\]
Note that this ratio completely specifies the posterior distribution of $\Theta_2$,
and hence the distribution of the utilities and all future computations.  Thus, whether
it is optimal to continue is a function only of this ratio, and thus of $s-f$.
%% In particular,
%% note that after equal numbers of successes and failures, $s-f=0$ and so the posterior 
%% distribution over $\Theta_2$ is equal to the prior.
For sufficiently low cost, the 
optimal policy samples when $s-f$ equals $-1$, $0$, or $1$.  But with probability
$1/3$, a state with $s-f=0$ transitions to another state $s-f=0$ after two samples, 
giving finite, although exponentially decreasing, probability to arbitrarily long 
sequences of computations.
\end{example}

However, in a number of settings, including the original Bernoulli model of \exampleref{example:bernoulli},
we can prove an upper bound on the number of computations.  For reasons of space,
and for its later use in \secref{sec:blinkered}, we prove here the bound for the one-armed Bernoulli model.

Before we can do this, we need to get an analytical handle on the optimal policy.
The key is through a natural approximate policy:

\begin{dfn}\label{dfn:myopic}
	Given a metalevel decision problem $M=(S,s_0,A_s,T,R)$,
	the \term{myopic policy} $\pi^m(s)$ is defined to equal $\argmax_{a\in A_s} Q^m(s,a)$ 
	where $Q^m(s,\bot) = \max_i \mu_i(s)$ and
	\begin{equation*}%% \label{eq:myopic}
		 Q^m(s,E) = \IE_M[ -c + \max_i \mu_i(S_1) \given S_0 = s, A_0 = E].		
	\end{equation*}
\end{dfn}

The myopic policy (known the metalevel greedy approximation with single-step
assumption in \cite{Russell+Wefald:1991a}) takes the best action, to either stop or perform a computation,
under the assumption that at most one further computation can be performed.
It has a tendency to stop too early, because changing one's mind about which real action to take often takes more than one computation.
In fact, we have:

%% \note{Theorem: if Optimal stops in x, myopic stops in x (converse is more useful!)}
	
\begin{thm}\label{thm:optimal-myopic}
	Given a metalevel decision problem $M=(S,s_0,A_s,T,R)$
	if the myopic policy performs some computation in state $s\in S$,
	then the optimal policy does too, i.e., if $\pi^m(s)\neq\bot$ then $\pi^*(s)\neq\bot$.
\end{thm}

\begin{hiddenproof}
	\begin{proof}
		Observe that the myopic Q-function \eqref{dfn:myopic} is equivalently given by
		\[
			Q^m(s,a) = Q^\bot(s,a)
		\]
		where $\bot$ is the policy which immediately stops $\bot(s)=\bot$.
		Thus $Q^m(s,a) \le Q^*(s,a)$.  If the optimal policy stops in a state $s\in S$ then
		\[
			Q^{\pi^*}(s,a) \le \max_i \mu_i(s),
		\]
		and so the same holds for $Q^m$, showing the myopic stops.
	\end{proof}
\end{hiddenproof}

% Similarity to an optimal stopping result
	% Search <blah> for one-step lookahead rules.
	% But HERE is generalized to a stopping problem where there are many ways to continue.

Despite this property, the stopping behavior of the myopic policy does
have a close connection to that of the optimal policy:

\begin{dfn}\label{dfn:closed}
	Given a metalevel decision problem $M=(S,s_0,A_s,T,R)$,
	a subset $S'\subseteq S$ of states is \term{closed under transitions}
	if whenever $s'\in S'$, $a\in A_{s'}$, $s''\in S$, and $T(s',a,s'')>0$,
	we have $s''\in S'$.
\end{dfn}

\begin{thm}\label{thm:myopic-optimal}
	Given a metalevel decision problem $M=(S,s_0,A_s,T,R)$
	and a subset $S'\subseteq S$ of states closed under transitions,	
	if the myopic policy stops in all states $s'\in S'$
	then the optimal policy does too.	
\end{thm}

\begin{hiddenproof}
	\begin{proof}
	Take any $s^*\in S'$, and note that all states the chain can transition
	to from $s^*$ are also in $S'$, by transition closure.  Defining $m(s) = \max_i\mu_i(s)$, 
	observe the myopic stopping for all such states implies that
	\begin{align*}
		\IE^{\pi}[(m(S_{j+1}) - c)\, 1(j<N)\given S_0=s^*] \\
		\le \IE^{\pi}[m(S_{j})\, 1(j<N)\given S_0=s^*]
	\end{align*}
	holds for all $j$, and as a result:
	\begin{align*}
		V^\pi(s) 
		&= \IE^{\pi}[ - c N + m(S_N) \given S_0=s^*] \\
		&= \IE^{\pi}[m(S_0) + \sum_{j=0}^{N-1} (m(S_{j+1}) - c - m(S_j)) \given S_0=s^*] \\
	%%	&\le \IE^{\pi}[m(S_0) + \sum_{j=0}^{N-1} 0 |S_0=s] \\		
		&\le \max_i\mu_i(s^*) \qedhere
	\end{align*}
	\end{proof}	
\end{hiddenproof}

Using these results connecting the behavior of the optimal and myopic policies, we can prove our bound:

\begin{thm}\label{thm:one-action-bound}
	The one-armed Bernoulli decision process with constant arm $\lambda\in[0,1]$ 
	performs at most $\lambda(1-\lambda)/c-3 \le 1/4c-3$ computations.
\end{thm}
\begin{proof}
Using \dfnref{dfn:myopic} and \exampleref{example:bernoulli2}, we determine
which states the myopic policy stops in by bounding $Q^m(s,E)$.  For a state $(s,f)$,
let $\mu=(s+1)/(n+2)$ be the mean utility for arm~2, where $n=s+f$.
Fixing $n$ and maximizing over $\mu$, we get sufficient condition for stopping
Since the set of states satisfying \eqref{eq:myopic-stopping} is closed under
\begin{equation}
	c \ge \frac{\lambda(1-\lambda)}{(n+3)} \qquad\qquad   n\ge \frac{\lambda(1-\lambda)}{c} - 3  \label{eq:myopic-stopping}
\end{equation}
Since the set of states satisfying \eqref{eq:myopic-stopping} is closed under
transitions ($n$ only increases), by \thmref{thm:optimal-myopic}.  Finally, note $\max_{\lambda\in[0,1]} \lambda(1-\lambda)=1/4$.
\end{proof}	

\begin{hiddenproof}
	\begin{proof}
		By \dfnref{dfn:myopic} and \exampleref{example:bernoulli2}, the myopic policy stops in
		a state $(s,f)$ when
		\begin{align}
			c \ge &\mu\max(\mu^+,m) + (1-\mu_i)\max(\mu^-, m) - \max(\mu,m) \label{eq:stopping}
		\end{align}
		where $\mu=(s+1)/(n+2)$  is the mean utility for arm~2, where $n=s+f$,
		$\mu^- = \mu - \mu/(n+3)$, and $\mu^+ = \mu + (1-\mu)/(n+3)$
		are the posterior means of arm~2 after simulating a failure and a success,
		%% and $\mu^-_i=(s_i+1)/(s_i+f_i+3) $ and $\mu^+_i = (s_i+2)/(s_i+f_i+3)$
		%% are the posterior means of action $i$ after simulating a failure and a success,
		respectively.  Whenever \eqref{eq:stopping} holds, stopping is preferred to sampling.

		Fixing $n$ and maximizing over $\mu$, we get sufficient condition for stopping
		\begin{equation}
			c \ge \frac{\lambda(1-\lambda)}{(n+3)} \qquad\qquad   n\ge \frac{\lambda(1-\lambda)}{c} - 3  \label{eq:myopic-stopping}
		\end{equation}
		Since the set of states satisfying \eqref{eq:myopic-stopping} is closed under
		transitions ($n$ only increases), by \thmref{thm:optimal-myopic}.  Finally, note $\max_{\lambda\in[0,1]} \lambda(1-\lambda)=1/4$.
	\end{proof}	
\end{hiddenproof}

A key implication is that the \emph{optimal} policy can be computed
in time $O(1/c^2)$, i.e., quadratic in the inverse cost.  This is particularly appropriate when the cost of 
computation is relatively high, such as in simulation experiments \citep{Swisher+et+al:2003},
or when the decision to be made is critical. 

%% Finite bounds for the $k$-armed problem
%% can also be derived and will be included in the full paper.

%% file: context.tex
%% \section{Context effects and non-indexability}

%% \note{No index theorem; via context inversion counter-example}

The Gittins index theorem \citep{Gittins:1979} is a famous structural
result for bandit problems.  It states that in bandit problems with
independent reward distribution for each arm and geometric discounting,
the optimal policy is an \term{index policy}:  each arm is assigned a
real-valued index based on its state only, such that it is optimal to
sample the arm with greatest index.

\begin{figure}[htb]
\centering
\includegraphics[scale=0.7, trim=90 70 400 300]{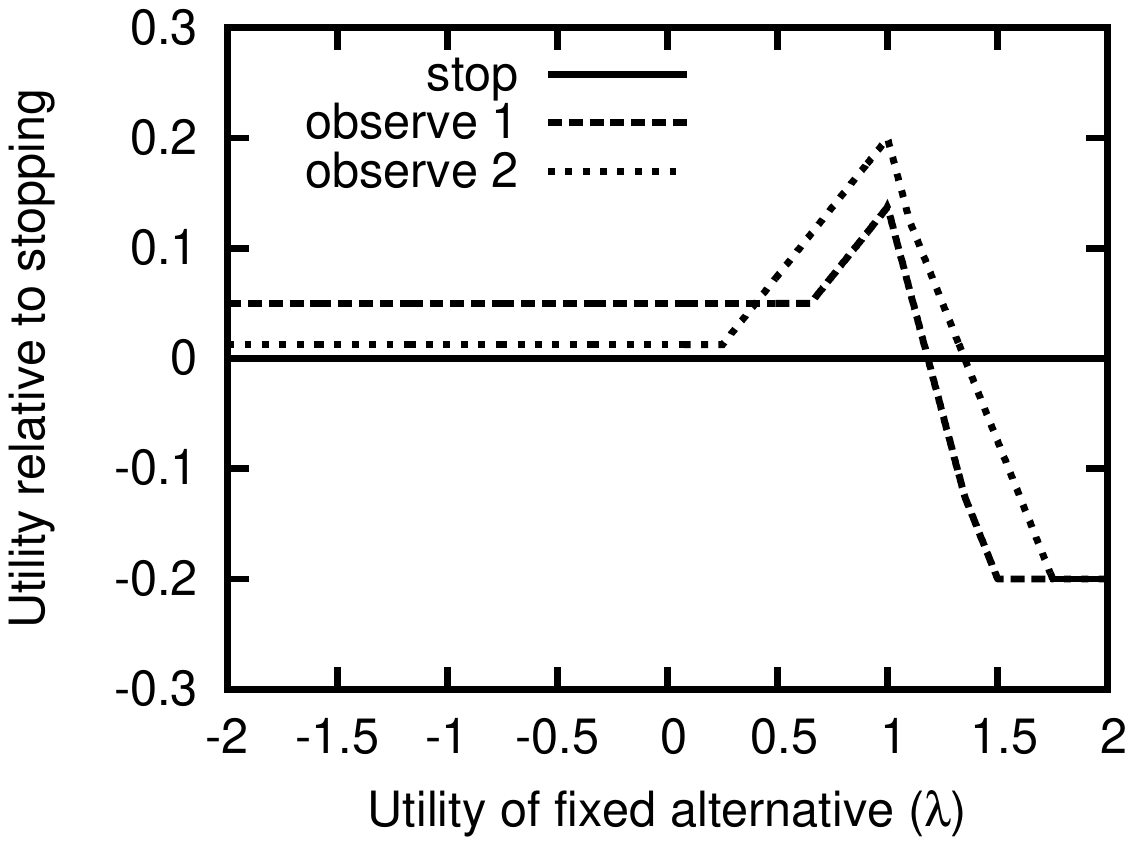}
\caption{Optimal Q-values of computing relative to stopping as a function of the utility of the fixed alternative.  Note the inversion
where for low $\lambda$ observing action~$1$ is strictly optimal, while for medium $\lambda$ observing action~$2$ is strictly optimal.}
\label{fig:swap-counterex}
\end{figure}

The analogous result does \emph{not} hold for metalevel decision problems,
even when the action's values are independent (this formalized later in \dfnref{dfn:independent-actions}):

\begin{example}[Non-indexability]
Consider a metalevel probability model with three actions. 
	$U_1$ is equally likely to be $-1.5$ or $1.5$ (low mean, high variance),
	$U_2$ is equally likely to be $0.25$ or $1.75$ (high mean, low variance),
	and $U_3=\lambda$ has a known value (the context).
The two computations are to observe exactly $U_1$ and $U_2$, respectively, each with cost $0.2$.
The corresponding metalevel MDP has 9 states and can be solved exactly.
\figref{fig:swap-counterex} plots $Q^*_\lambda(s_0,U_i) - Q^*_\lambda(s_0,\bot)$ as a function of the known value $\lambda$.
As the context $\lambda$ varies the optimal action inverts from observing 1 to observing 2.
Inversions like this are impossible for index policies.
\end{example}

\begin{comment}
	set samples 1000

	max(x,y) = x>y ? x : y
	max3(x,y,z) = max(max(x,y),z)

	q0(u) = max(1,u)
	qa(u) = 0.5*max3(u, 1.5, -0.2 + 0.5*max(u,1.5) + 0.5*max(u, 1.75)) + 0.5*max3(u, 1, -0.2 + 0.5*max(u,0.25) + 0.5*max(u, 1.75)) - 0.2
	qb(u) = 0.5*max3(u, 0.25, -0.2 + 0.5*max(u,1.5) + 0.5*max(u, 0.25)) + 0.5*max3(u, 1.75, -0.2 + 0.5*max(u,1.75) + 0.5*max(u, 1.75)) - 0.2

	set terminal postscript enhanced linewidth 2
	set size 0.5,0.5
	set output "swap-counterex.ps"
	set xlabel "Utility of fixed alternative ({/Symbol l})"
	set ylabel "Utility relative to stopping"
	set xtics 0.5
	set ytics 0.1
	set key left top
	plot [u=-2:2] [-0.3:0.3] 0 title "stop" with lines lw 2, qa(u)-q0(u) title "observe 1" with lines lw 2, qb(u)-q0(u) title "observe 2" with lines lw 2	
\end{comment}

%% \note{Theorem: /stopping/ context effect occurs only for a single convex interval of
%%    context value (how general can we make this?)}

There is, however, a restriction on what kind of influence the context can have:

\begin{dfn}\label{dfn:context}
	Given a metalevel decision problem $M=(S,s_0,A_s,T,R)$, and a constant $\lambda\in\R$,
	define $M_\lambda = (S,s_0,A_s,T,R_\lambda)$ to be $M$ with an additional action of known 
	value $\lambda$, defined by:
	\begin{align*}
		R_\lambda(s,E,s')      &= R(s,E,s') \\
		R_\lambda(s,\bot,\bot) &= \max\{\lambda, R(s,\bot,\bot)\}
	\end{align*}
\end{dfn}

Note this is equivalent to adding an extra arm with constant value $U_{k+1}=\lambda$.

\begin{thm}
	Given a metalevel decision problem $M=(S,s_0,A_s,T,R)$, 
	there exists a real interval $I(s)$ for every state $s\in S$ such that
	it is optimal to stop in state $s$ in $M_\mu$ iff $\mu\notin I(s)$.
	Furthermore, $I(s)$ contains $\max_i\mu_i(s)$ whenever it is nonempty.
\end{thm}

\begin{hiddenproof}
	\begin{proof}
	With $m(s) = \max_i\mu_i(s)$, the utility of a policy $\pi$ starting in state $s$ of $M_\mu$ is
	\[
		V^\pi_{M_\mu}(s) = \IE_{\pi}[-c\,N + \max(\mu,m(S_N)) \given S_0=s]
	\]
	and the utility of stopping in this state $\max(\mu,m(s_0))$.
	We wish to show that the set of $\mu$ such that
	\[
		\max_\pi \IE_{\pi}[-c\,N + \max(\mu,m(S_N)) - \max(\mu,m(S_0)) \given S_0=s] \le 0
	\]
	forms an interval.  

	Observe that for any random variable $X$, 
		$\IE[\max(\mu,X)]$ is monotonically increasing in $\mu$ with subderivative between zero and one.
	As a result, for any $v_1$
		$\IE[\max(\mu,X)] - \max(\mu,v_1)$ 
		is monotonically increasing for $\mu<v$, 
		and monotonically decreasing thereafter.
	Therefore, the set if $\mu$ such that this expression is at most $v_2$ forms an interval, containing $v_1$ if non-empty.

	Applying this with $v_1 = m(s_0)$ and $v_2=\IE_{\pi}[c\,N]$, and observing that the union of intervals containing
	a point is an interval containing that point, gives the result.
	\end{proof}	
\end{hiddenproof}

%% \note{Example showing that choice between computations to perform 
%% can vary arbitrarily many times?  Does one exist?}

%% file: blinkered.tex
%% \section{The blinkered policy}

%% \section{The blinkered policy}

%% This is the key section.  Write it, and then see what prereqs are required.

The myopic policy is an extreme approximation, often stopping far too early.
A better approximation can be obtained, at least for the case where
each computation can only affect the value of one action.
The technical definition (closely related to {\em subtree independence} in 
Russell and Wefald's work) is as follows:

\begin{dfn}\label{dfn:independent-actions}
	A metalevel probability model $(U_1,\dots,U_k,\Evidence)$ 
	has \term{independent actions} if the computational variables can be partitioned 
	$\Evidence = \Evidence_1\cup\dots\cup\Evidence_k$ such that such that
	the sets $\{U_i\}\cup\Evidence_i$ are independent of each other for different $i$.	
\end{dfn}

With independent actions, we can talk about metalevel policies that focus on
computations affecting a single action. These policies are not myopic---they can consider arbitrarily many computations---but they are {\em blinkered} because they can look
in only a single direction at a time:

\begin{dfn}\label{dfn:blinkered}
	Given a metalevel decision problem $M=(S,s_0,A_s,T,R)$ with independent actions,
	the \term{blinkered policy} $\pi^b$ is defined by $\pi^b(s) = \argmax_{a\in A_s} Q^b(s,a)$ where
	$Q^b(s,\bot) = \bot$ and for $E_i\in\Evidence_i$
	\begin{equation}\label{eq:blinkered}
		Q^b(s,E_i) = \sup_{\pi\in\Pi^b_i} Q^\pi(s,E_i)
	\end{equation}
	where $\Pi^b_i$ is the set of policies $\pi$ where $\pi(s)\in\Evidence_i$ for all $s\in S$.
\end{dfn}

Clearly, blinkered policies are better than myopic: $Q^m(s,a) \le Q^b(s,a) \le Q^*(s,a)$.
Moreover, the blinkered policy can be computed in time proportional to the number of arms, by breaking the
decision problem into separate subproblems:

\begin{dfn}\label{dfn:one-action}
	Given a metalevel decision problem $M=(S,s_0,A_s,T,R)$ with independent actions,
	a \term{one-action metalevel decision problem} for $i=1,\dots,k$ is the metalevel decision
	problem $M^1_{i,\lambda} = (S_i,s_0,A_{s0},T_i,R_i)$ defined by the metalevel probability
	model $(U_0,U_i,\Evidence_i)$ with $U_0=\lambda$.
\end{dfn}

Note that given a state $s$ of a metalevel decision problem, we can form a state
$s_i$ by taking only the results of computations in $\Evidence_i$ (see \dfnref{dfn:metalevel-mdp}).
By action independence, $\mu_i(s)$ is a function only of $s_i$.

\begin{thm}\label{thm:blinkered}
Given a metalevel decision problem $M=(S,s_0,A_s,T,R)$ with independent actions,
let $M^1_{i,\lambda_i}$ be the $i$th one-action metalevel decision problem for $i=1,\dots,k$.
Then for any $s\in S$, whenever $E_i\in A_s\cap\Evidence_i$ we have:
\[
	Q^b_M(s,E_i) = Q^*_{M^1_{i,\mu^*_{-i}}}(s_i, E_i)
\]
where $\mu^*_{-i} = \max_{j\neq i} \mu_j(s)$.
\end{thm}

\begin{hiddenproof}
	\begin{proof}
	Fix a state $s$, a $E_i\in A_s$ and take any $\pi\in\Pi^b_i$.  Note that such policies
	are equivalent to polices $\pi'$ on $M^1_{1,m}$, and all such policies are represented.
	Consider $Q^\pi(s,E_i)$.  As $\pi(s)\in\Evidence_i$ for all $s\in S$, by action independence $\mu_j(S_n) = \mu_j(s)$.
	By this and \thmref{thm:value-of-computation}, then,
	\[
		Q^\pi_M(s,E_i) = \IE^\pi_M[ -c\,N + \max(\mu_i(S_N), m_i) \given S_0=s, A_0=E_i].
	\]
	Noting that $\mu_i(S_N)$ is a function only of $(S_N)_i$, and that since 
	But then this is exactly $Q^*_{M^1_{i,\mu^*_{-i}}}(s_i, E_i)$.  Taking the supremum
	over $\pi$ gives the result.
	\end{proof}	
\end{hiddenproof}

\thmref{thm:blinkered} shows that to compute the blinkered policy we need
only compute the optimal policies for $k$ separate one-action problems.

For the Bernoulli problem with $k$ actions, the one-action metalevel decision problems
are all one-action Bernoulli problems (\exampleref{example:bernoulli}).  By \thmref{thm:one-action-bound}
these policies perform at most $1/4c - 3$ computations.
As a result, the blinkered policy can be numerically computed in time $O(D/c^2)$ 
independent of $k$ by backwards induction, where $D$ is the number of points $\lambda\in[0,1]$
for which we compute $Q^*_{M^1_{i,m}}(s)$.\footnote{in our experiments below, $D=129$ points are equally spaced,
using linear interpolation between points.}  This will be worth the cost in 
the same situations as mentioned at the end of \secref{sec:optimal}.

\begin{figure}[htb]
\centering
\includegraphics[scale=0.7, trim=90 70 400 300]{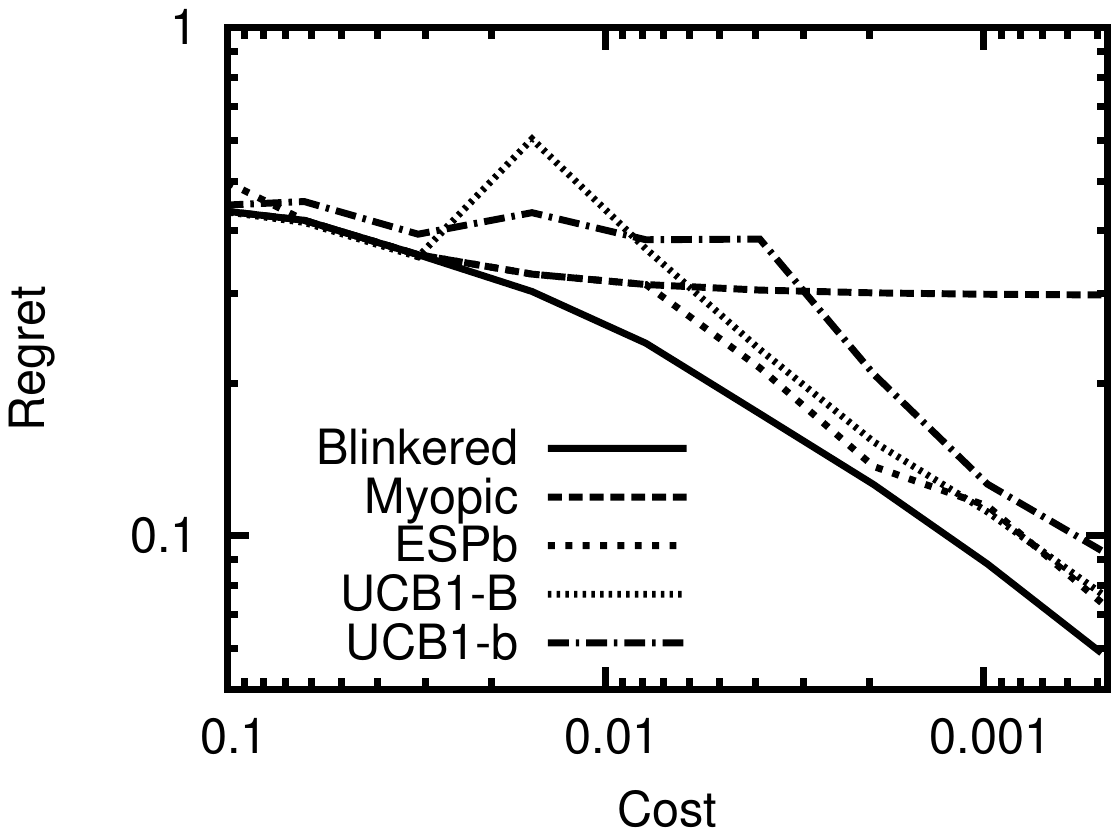}
\caption{Average regret of various policies as a function of the cost
in a 25-action Bernoulli sampling problem, over 1000 trials. Error bars
omitted as they are negligible (the relative error is at most 0.03).}
\label{fig:blinkered}
\end{figure}

\figref{fig:blinkered} compares the blinkered policy 
to several other policies from the literature, using a
Bernoulli sampling problem with $k=25$ and a wide range of values for the step cost $c$.
Performance is measured by expected {\em regret}, where the regret includes the cost of sampling:
$R = (\max_i U_i) - U_j + c\,n$
where $n$ is the number of computations and $j$ is the action actually selected.
The blinkered policy significantly outperforms all others.  The myopic policy
plateaus as it quickly reaches a position where no single computation can
change the final action choice.  ESPb performs quite well given
that is making a normal approximation to the Beta posterior.  
The curves for UCB1-B and UCB1-b show that even given a good stopping rule, UCB1's
choice of actions to sample is not ideal.

%% file: conclusion.tex
\section{Conclusion}

The selection problem has numerous applications. This paper formalized the
problem as a belief-state MDP and proved some important properties of the 
resulting formalism. An application of the selection problem to control of
sampling was examined, and the insights provided by properties of the MDP 
led to approximate solutions that improve the state of the art. This was
shown in empirical evaluation both in ``flat" selection and when extending
the methods to game-tree search for the game of Go.

The methods proposed in the paper open up several new research
directions. The first is a better approximate solution of the MDP,
that should lead to even better flat sampling algorithms for
selection. A more ambitious goal is extending the formalism to
trees---in particular, achieving better sampling at non-root nodes,
for which the purpose of sampling differs from that at the root.

%% file: acknowledgements.tex
\section*{Acknowledgments}

The research is partially supported by Israel Science Foundation grant 305/09,
National Science Foundation grant IIS-0904672,
DARPA DSO FA8650-11-1-7153,
the Lynne and William Frankel Center for Computer Sciences,
and by the Paul Ivanier Center for Robotics Research and Production Management.